\pdfoutput=1

\documentclass[11pt]{article}

\usepackage{EMNLP2022}

\usepackage{times}
\usepackage{latexsym}
\usepackage{times}
\usepackage{latexsym}
\usepackage{hyperref}       
\usepackage{url}            
\usepackage{booktabs}       
\usepackage{amsfonts}       
\usepackage{nicefrac}       
\usepackage{microtype}      
\usepackage{lipsum}
\usepackage{makecell}
\usepackage{graphicx}
\graphicspath{ {./images/} }
\usepackage{here}
\usepackage{stmaryrd}
\usepackage{multirow}
\usepackage{enumitem}
\usepackage{subfig}
\usepackage{dblfloatfix}
\usepackage{changepage}
\usepackage{threeparttable}
\usepackage{tabu}
\usepackage{longtable}
\usepackage{float}
\usepackage{amsmath}
\usepackage{booktabs}
\usepackage{tabularx}
\usepackage{longtable}

\usepackage{array}

\usepackage[T1]{fontenc}

\usepackage[utf8]{inputenc}

\usepackage{microtype}

\usepackage{inconsolata}

\title{Explainable Verbal Deception Detection using Transformers}

\author{Loukas Ilias\textsuperscript{1}, Felix Soldner\textsuperscript{2,3}, and Bennett Kleinberg\textsuperscript{3,4} \\
        \\
        \textsuperscript{1}Decision Support Systems Laboratory, School of Electrical and Computer Engineering, \\
        National Technical University of Athens, 15780 Athens, Greece \\
        \textsuperscript{2}GESIS – Leibniz Institute for the Social Science, Köln, Germany\\ 
        \textsuperscript{3}Department of Security and Crime Science \& Dawes Centre for Future Crime, \\
        University College London \\
        \textsuperscript{4}Department of Methodology and Statistics, Tilburg University \\
        \texttt{lilias@epu.ntua.gr}, \\
        \texttt{felix.soldner@gesis.org}, \\
        \texttt{bennett.kleinberg@tilburguniversity.edu}
        }

\begin{document}
\maketitle
\begin{abstract}
People are regularly confronted with potentially deceptive statements (e.g., fake news, misleading product reviews, or lies about activities). Only few works on automated text-based deception detection have exploited the potential of deep learning approaches. A critique of deep-learning methods is their lack of interpretability, preventing us from understanding the underlying (linguistic) mechanisms involved in deception. However, recent advancements have made it possible to explain some aspects of such models. This paper proposes and evaluates six deep-learning models, including combinations of BERT (and RoBERTa), MultiHead Attention, co-attentions, and transformers. To understand how the models reach their decisions, we then examine the model's predictions with LIME. We then zoom in on vocabulary uniqueness and the correlation of LIWC categories with the outcome class (truthful vs deceptive). The findings suggest that our transformer-based models can enhance automated deception detection performances (+2.11\% in accuracy) and show significant differences pertinent to the usage of LIWC features in truthful and deceptive statements.
\end{abstract}

\section{Introduction}
Computer-automated verbal deception detection has gained much attention in recent years \citep{kleinberg2018using,KHAN2021114341,chebbi2021deception}. During this time, studies examining deception detection have used data from various domains including news \citep{rubin2015deception,conroy2015automatic,girgis2018deep}, customer reviews \citep{kennedy-etal-2019-fact,hajek2020fake,8259828,ott-etal-2011-finding}, court testimonies \citep{fornaciari-poesio-2012-decour,fornaciari2013automatic,fornaciari-etal-2021-bertective,kao2020detecting}, or interactions in lie games \citep{soldner-etal-2019-box}. Other work has collected text data from experiments about topics such as personal/political opinions \citep{capuozzo-etal-2020-decop}, people's weekend plans \citep{KLEINBERG2021103250} or intentions about flying \citep{kleinberg2017investigation}. 

For the task of deception detection, it is worth demarcating deception from related tasks, particularly those of fact-checking and misinformation detection. Our work is related to efforts on interpretable fact-checking (e.g., \citep{Nguyen_Kharosekar_Lease_Wallace_2018}) in that it aims to understand how classifiers use the input information, but it differs in the type of deception. The core difference between deception (lie) detection and fact-checking or misinformation detection is that deception detection often does not lend itself to external verification of truth absent additional information. For example, someone lying about their activities during the last weekend may state that they watched a movie at home. This statement cannot easily be submitted to external verification queries as is possible in fact-checking or in some cases even misinformation detection (if external information is available). As such, deception detection research is particularly eager to use linguistic patterns as a means to identify truths or lies (see \citet{fitzpatrick2015automatic}). More so, even within deception detection research - as defined above - two prominent types of lies are differentiated (see \citet{nahari2019language}): lies of commission are those where a liar makes a false statement (e.g., about their weekend plans), while lies of omission are an attempt to hide events or facts that did happen (e.g., that someone visited a bar yesterday). For a comprehensive review of challenges for computational approaches to deception detection from language, see \citet{tomas2022computational}. Moreover, for a review on general pitfalls and opportunities in verbal deception detection from a theoretical and psychological angle, see \citet{vrij2010pitfalls}.

However, most approaches in deception detection use only traditional (ensemble) machine learning methods, such as logistic regression, support vector machines, random forests, or naive bayes classifiers to predict veracity \citep{Wu_Singh_Davis_Subrahmanian_2018,10.1145/2701126.2701130,fornaciari2013automatic}. 
Only few studies have explored deep learning methods based on transformer-networks \citep{fornaciari-etal-2021-bertective,kao2020detecting,kennedy-etal-2019-fact}, which show increased prediction performances over the traditional approaches, but do not focus on interpreting the linguistic properties of the texts. Since previous research has found mostly mixed results about deceptive cues in text, it is important to be able to interpret more powerful methods, to add to our understanding of deception as expressed in text. However, deep learning models are difficult to interpret and often called black boxes, due to their high complexity of many layers involving a large number of (combinations of) trainable parameters.

Recently, interpreting deep learning methods has become more feasible and can be described as \textbf{(1)} local post-hoc (i.e. the explanation of a single prediction by performing additional operations), \textbf{(2)} local self-explaining (i.e. the explanation of a single prediction using the model itself), \textbf{(3)} global post-hoc (i.e. by performing additional operations for explaining the entire model’s predictive reasoning) and \textbf{(4)} global self-explaining (i.e. using the predictive model itself for explaining its entire predictive reasoning) \citep{danilevsky-etal-2020-survey}. Several explainability techniques belonging to the aforementioned categories have been introduced including the visualization of neural networks \citep{simonyan2014deep,li2015visualizing}, so-called surrogate models \citep{ribeiro2016should}, and layer-wise relevance propagation \citep{binder2016layer}.
However, for the task of deception detection, researchers usually face the dilemma of choosing between high model performance promises from deep learning models or easier interpretability from simpler approaches. Almost no prior work aimed to explore explainability techniques for interpreting the predictions made by deep learning architectures for deception detection. Thus, with this paper we aim to add to the understanding of deceptive text characteristics by implementing several deep learning approaches and interpreting the best performing method. Our main contributions are:

\begin{itemize}
    \item trying to detect deception using two statements concurrently inspired by the success of the siamese neural networks.
    \item comparing transformer-based networks to shallow machine learning algorithms and human baseline approaches.
    \item performing an analysis of the linguistic patterns of truthful and deceptive texts.
    \item employing LIME \citep{ribeiro2016should} to examine the predictions made by the best performing model. 
\end{itemize}

\section{Related Work} \label{relatedwork}

A number of studies have proposed systems to address automated verbal deception detection. Some works made datasets publicly available, some of which were transcripts of court cases  \citep{fornaciari-poesio-2012-decour,KLEINBERG2021103250,perez-rosas-etal-2015-verbal}, were generated in the laboratory \citep{ott-etal-2011-finding} or stem from transcripts of lie games \citep{soldner-etal-2019-box}. More recently, \citet{fornaciari-etal-2021-bertective} used the data by \citet{fornaciari-poesio-2012-decour} and introduced contextualized and non-contextualized transformer-based networks for detecting deception from real-life court transcripts. The authors found that no model significantly outperformed the accuracy obtained by the SVM classifier \citep{fornaciari2013automatic}. For finding the linguistic patterns used in deceptive language, the authors looked at Information gain of \textit{n}-grams \citep{forman2003extensive} and the sampling and occlusion algorithm \citep{2019arXiv191106194J}. They concluded that deceptive statements contain more negations, and were less assertive in contrast to truthful statements.

Another deception detection study proposed a neural network architecture consisting of a fully-connected layer and/or an LSTM layer \citep{kao2020detecting}. They showed that one-hot-encoding yields the highest F1-score compared to embedding models obtained via fastText and BERT \citep{devlin2019bert}. Another task of verbal deception detection is fake review detection \citep{kennedy-etal-2019-fact,hajek2020fake,8259828,ott-etal-2011-finding}. Findings in this area suggest that the use of transformer-based networks, e.g., BERT, may improve detection results. 


\section{Datasets} \label{Dataset_Kleinberg}

\subsection{Deception in future activities} \label{kleinberg_dataset}
For our experiments, we use the publicly available dataset introduced by \citet{KLEINBERG2021103250}. The data were collected through crowdsourcing (via Prolific Academic) in which participants were asked to provide a statement about their most significant non-work-related activity in the next seven days. Participants were asked the two following questions:
\begin{itemize}
    \item \textbf{Q1:} “Please describe your activity as specific as possible” 
    \item \textbf{Q2:} “Which information can you give us to reassure us that you are telling the truth”. 
\end{itemize} 
Each participant was randomly assigned to either the truthful condition (i.e. they answered the questions about an activity they genuinely planned to do) or the deceptive condition. In the deceptive condition, participants were assigned matched activities from the truthful condition.\footnote{The matching ensured that the participants were given an activity they had not planned themselves so that fully deceptive statements were written.} 

The dataset contains 1640 statements (857 deceptive and 783 truthful) consisting of two answers per participant.

We opted for this dataset, since it can be characterized as of high quality due to the strong experimental control during the collection procedure. Thus, the ground truth about each statement is known, which is often uncertain in other datasets.

\subsection{Open Domain Deception}\label{Dataset_Open}
For showing the robustness of our proposed approaches towards the task of deception detection we use an additional dataset: the Open Domain Deception Dataset \citep{perez-rosas-mihalcea-2015-experiments}. Here, the authors asked 512 participants through Amazon Mechanical Turk to provide seven truths and seven lies, on topics of their own choice. For conducting our experiments, we merged statements of each participant into one. Thus, the dataset includes 512 truthful and 512 deceptive statements.

\section{Predicting Deceptive Statements} \label{methods}





Here we detail the prediction models that we compare on the abovementioned datasets.

\textbf{BERT+Dense Layers.} The BERT model \citep{devlin2019bert} receives as input the answer to the question \textbf{Q1}\footnote{For the Open Domain Deception Dataset, the data provided by each participant serves analogously as Q1.}. The sequence of hidden-states at the output of the model's last layer constitutes the output of the BERT model. That output is passed through a mean-pooling layer whose output is passed through three dense layers to obtain the final output.

\noindent \textbf{BERT+MultiHead Attention.} The approach is identical to the one above, with the addition that the output of the BERT model is passed through a multi-head attention layer consisting of six heads. Finally, the output of the transformer is fed to three dense layers, which produce the final prediction.

\noindent \textbf{BERT+Transformer.} Identical to \textbf{BERT+Dense Layers} with the difference that the output of the BERT model is passed through the encoder layer of the transformer \citep{NIPS2017_3f5ee243} consisting of six layers and six heads. Finally, the output of the transformer is fed to a mean pooling layer. The output of the pooling layer is fed to three ReLU activated Dense layers of size 512, 128, 64 and finally to a dense layer with a sigmoid activation function and one unit, which results in the final prediction.

\noindent \textbf{BERT+Co-Attention.} The model receives the two answers to questions \textbf{Q1} \& \textbf{Q2} as input.\footnote{In terms of the Open Domain Deception Dataset, we split each statement into two statements of equal length.} Both answers are fed to two siamese BERT models (sharing the same weights). Next, we employ the co-attention mechanism introduced by \citet{NIPS2016_9dcb88e0} over the two embeddings (of the two statements) in an attempt to improve detection performance. 

Let $C \in \mathbb{R}^{d \times N}$ and $S \in \mathbb{R}^{d \times T}$ be the outputs of the BERT model, where  $d$ indicates the hidden size of the BERT model and is equal to 768, while \textit{N} \& \textit{T} constitute the number of tokens in answers to the questions \textbf{Q1} \& \textbf{Q2} respectively. We have omitted the first dimension, which corresponds to the batch size. Following the methodology proposed by \citet{NIPS2016_9dcb88e0}, firstly we compute the affinity matrix $F \in \mathbb{R}^{N \times T}$ as:

\begin{equation}
    F = \tanh(C^T W_l S)
    \label{equation1}
\end{equation}

\noindent where $W_l \in \mathbb{R}^{d \times d}$ contains the weights. Next, by considering this affinity matrix as a feature, we learn to predict the attention maps for both statements via the following,

\begin{equation}
\begin{aligned}
    H^s = \tanh(W_s S + (W_c C)F)
    \\
    H^c = \tanh(W_c C + (W_s S)F^T)
    \end{aligned}
\end{equation}

\noindent where $W_s, W_c \in \mathbb{R}^{k \times d}$ are the weight parameters. The attention probabilities for each word in both statements are calculated via the following,

\begin{equation}
\begin{aligned}
    a^s = softmax(w_{hs} ^T H^s) \\
    a^c = softmax(w_{hc} ^T H^c)
    \end{aligned}
\end{equation}

\noindent where $a_s \in \mathbb{R}^{1 \times T}$ and $a_c \in \mathbb{R}^{1 \times N}$. $W_{hs}, W_{hc} \in \mathbb{R}^{k \times 1}$ are the weight parameters. Based on the above attention weights, the attention vectors for each statement are calculated as the weighted sum of the features from each statement. Formally,
\begin{equation}
    \hat{s} = \sum_{i=1}^{T} a_i ^s s^i, \quad \hat{c}=\sum_{j=1}^{N} a_j ^c c^j
\end{equation}

\noindent where $\hat{s} \in \mathbb{R}^{1 \times d}$ and $\hat{c} \in \mathbb{R}^{1 \times d}$.

Then, we concatenate these two vectors, so that
\begin{equation}
z=[\hat{s},\hat{c}]
\label{equation2}
\end{equation}

\noindent where $z \in \mathbb{R}^{1 \times 2d}$ and we feed the vector \textit{z} to a series of dense layers, to obtain the output.

\noindent \textbf{BERT+Co-Attention+LIWC.} Identical to \textbf{BERT+Co-Attention} with the difference that the output of the co-attention mechanism \textit{z} is fed to a Dense layer, where the ouput of this dense layer is concatenated with Linguistic Inquiry and Word Count (LIWC) features obtained by the statements-answers to question \textbf{Q1}. The LIWC software offers a (psycho)linguistic dictionary developed with domain experts and measures constructs such as cognitive processes, references to time, or personal concerns \citep{pennebaker2015development}. Numerous works at the intersection of psychology and computational linguistics have used the LIWC (e.g., \citep{Tausczik2010ThePM,perez-rosas-etal-2018-automatic}) and recently the newest version “LIWC-22” \citep{boyd2022development} has been released (note that we used the features obtained with the 2015 version). We feed the concatenated vector to two dense layers, so as to get the final prediction.

\noindent \textbf{BERT+Transformers+Co-Attention.} As above, the model receives the two answers to questions \textbf{Q1} \& \textbf{Q2} as input. Both answers are fed to two siamese BERT models (sharing the same weights). Each output of the BERT model is passed through the encoder layer of the transformer with six layers and six heads. Then, the two outputs of the respective transformers, namely $C$ \& $S$ are passed through a co-attention mechanism, as described via the equations \ref{equation1}-\ref{equation2}. Then the output of the co-attention mechanism \textit{z} is fed to a series of Dense layers, which produce the final output.

We experiment also with RoBERTa \citep{https://doi.org/10.48550/arxiv.1907.11692} instead of BERT and lower performance results are achieved. 

\section{Experiments} \label{experiments}

\textbf{Baselines.} Regarding the dataset by \citet{KLEINBERG2021103250}, we compare the performance of the proposed approaches introduced in Section \ref{methods} against the performing system adopted by \citet{KLEINBERG2021103250}, who used both machine learning and human judgements to classify veracity. The baselines are presented below.
\begin{itemize}[noitemsep]
    \item \textit{Human Baseline.} In this framework, the participants were asked to read each statement and indicate their judgement on a slider ranging from 0 to 100, with 0 meaning certainty truthful, whereas 100 meaning certainty deceptive.
    \item \textit{Machine Learning: LIWC.} This method trains a random forest classifier by using all 93 categories of the LIWC as the feature set.
    \item \textit{Machine Learning: POS.} This approach exploits relative part-of-speech (POS) frequencies. A random forest classifier is trained with the corresponding feature set.
\end{itemize}

Regarding the experiments conducted on the Open Domain Deception Dataset, we compare the performance of our introduced approaches with:
\begin{itemize}
    \item \textit{Feature Set: POS}: This method extracts POS tags per statement and trains an SVM Classifier.
\end{itemize} 
\noindent \textbf{Experimental Setup.} We use a 5-fold stratified cross-validation procedure with five repetitions for training and evaluating the introduced approaches. Within each loop we divide the training set into a training and evaluation set (80\% - 20\% respectively). We use the base-uncased version of BERT model from the Transformers library \citep{wolf-etal-2020-transformers}. Furthermore, the SGD optimizer is used for all architectures, except for \textit{BERT + Co-Attention + LIWC} where we use the Adam optimizer, and train the proposed architectures by minimizing the binary cross entropy loss function. The initial learning rate for all architectures is 0.001. For dealing with the imbalanced dataset, we introduce balanced class weights to the loss function. During the training an early stopping method with a patience of 10 is applied based on the validation loss. Concurrently, we apply the callback ReduceLROnPlateau, where we reduce the learning rate by a factor of 0.1 if the validation loss has not been decreased for three consecutive epochs. Specifically, we train the architectures by freezing the weights of the BERT model. After having trained the architectures, we unfreeze the weights of the BERT model and we keep training the proposed architectures, where an early stopping method with a patience of 2 is applied based on the validation loss for avoiding overfitting. All models are created using the Tensorflow library \citep{abadi2016tensorflow} and trained on a single 24Gb NVIDIA Titan RTX GPU.

\noindent \textbf{Evaluation Metrics.} We report the Accuracy, Precision, Recall, F1-score, Area Under the ROC Curve (AUROC), and Specificity as performance metrics. All these metrics were calculated by considering the deceptive statements as the positive class and the truthful statements as the negative class.

\section{Results} \label{results}

Table \ref{compare} presents the results of both the baselines and the introduced transformer-based models for the experiments on the dataset described in Section \ref{kleinberg_dataset}.\footnote{The results of RoBERTa for the dataset by \citet{KLEINBERG2021103250} are reported in Appendix~\ref{results_roberta}.} The results of the proposed models on the Open Domain Deception Dataset are reported in Appendix~\ref{open_domain_results}.

Regarding the experiments conducted on the dataset described in Section \ref{kleinberg_dataset}, we observe that in the previous traditional models, the \textit{Machine Learning: LIWC} approach shows best performances in most metrics (Precision, Recall, F1-score, Accuracy, and AUROC) and is used as a baseline for comparison with the introduced transformer-based models.

Of the new approaches, BERT+Co-Attention performs best in most metrics. Compared to the next best result (BERT+Co-Attention+LIWC and BERT+Transformers+Co-Attention), the model increased  precision by 0.18 points, recall by 1.12 points, the F1-score by 0.60 points, and Accuracy by 0.45 points. In terms of AUROC score, BERT+Co-Attention performs slightly worse than BERT+Co-Attention+LIWC (-0.06 points). Furthermore, one can observe that BERT+Co-Attention, BERT+Co-Attention+LIWC, and BERT+Transformers+Co-Attention yield better evaluation results than BERT+Dense Layers, BERT+MultiHead Attention, and BERT+Transformers, which indicates, that using both statements is useful towards a better classification performance of categorizing truthful and deceptive statements. Also, we observe that BERT+Co-Attention+LIWC performs slightly worse than BERT+Co-Attention. This slight difference is attributable to the usage of the concatenation operation (i.e., LIWC features are concatenated with the representation vector obtained from both statements). In addition, the difference in performance between BERT+Co-Attention and BERT+Transformers+Co-Attention is attributable to the usage of the encoder layer of the transformer. Specifically, the addition of the Transformer consisting of six layers and six heads may lead to overfitting and decrease of the performance due to the small dataset used.

It is worth mentioning that BERT+Co-Attention outperforms the previous results obtained by \textit{Machine Learning: LIWC}, on all evaluation metrics (Precision: +2.43 points, Recall: +1.72 points, F1-score:+1.71 points, Accuracy: +2.11 points, AUROC: +2.77 points, Specificity: +2.59 points).

\begin{table*}[hbt]
\scriptsize
\centering
\begin{tabular}{lcccccc}
\toprule
\multicolumn{1}{l}{}&\multicolumn{6}{c}{\textbf{Evaluation metrics}}\\
\cline{2-7} 
\multicolumn{1}{l}{\textbf{Architecture}}&\textbf{Precision}&\textbf{Recall}&\textbf{F1-score}&\textbf{Accuracy}&\textbf{AUROC}&\textbf{Specificity}\\
\hline
\textit{\small{Human Baseline}} & 54.66 & 24.40 & 33.74 & 50.15 & 52.00 & 78.06\\
\textit{\small{Machine Learning: LIWC}} & 67.71 & 76.02 & 71.62 & 68.50 & 75.00 & 60.26\\
\textit{\small{Machine Learning: POS}} & 63.68 & 70.76 & 67.03 & 63.61 & 67.00 & 55.77\\
\hline \hline
\textit{\small{BERT + Dense Layers}} & 68.90 & 65.92 & 66.67 & 66.36 & 73.09 & 66.90 \\
& $\pm$3.15 & $\pm$10.92 & $\pm$8.01 & $\pm$3.74 & $\pm$4.00 & $\pm$7.84\\ \hline
\textit{\small{BERT + MultiHead Attention}} & 69.86 & 71.64 & 70.65  & 68.89 & 76.43 & 65.85 \\
& $\pm$2.68 & $\pm$3.76 & $\pm$2.09 & $\pm$2.24 & $\pm$1.88 & $\pm$5.21\\ \hline
\textit{\small{BERT + Transformers}} & 69.89 & 72.68 & 71.16 & 69.33 & 77.18 & 65.68\\
& $\pm$2.59 & $\pm$3.93 & $\pm$2.05 & $\pm$1.88 & $\pm$1.81 & $\pm$4.61\\ \hline
\textit{\small{BERT + Co-Attention}}& \textbf{70.14} & \textbf{77.74} & \textbf{73.33} & \textbf{70.61} & 77.77 & 62.85  \\
& $\pm$4.39 & $\pm$7.76 & $\pm$2.59 & $\pm$2.58 & $\pm$2.88 & $\pm$10.25 \\ \hline
\textit{\small{BERT + Co-Attention + LIWC}} & 69.96 & 76.66 & 72.73 & 70.16 & \textbf{77.83} & 63.04  \\
& $\pm$3.76 & $\pm$8.22 & $\pm$2.71 & $\pm$1.90 & $\pm$2.53 & $\pm$9.37\\ \hline
\textit{\small{BERT + Transformers + Co-Attention}} & 69.96 & 74.19 & 71.94 & 69.77 & 77.16 & \textbf{64.93} \\
& $\pm$2.76 & $\pm$3.73 & $\pm$2.29 & $\pm$2.39 & $\pm$1.83 & $\pm$4.92\\ 
\bottomrule
\end{tabular}
\caption{Performance comparison between previous \citep{KLEINBERG2021103250} and proposed deception detection models. Reported values are means $\pm$ standard deviation. Best results per evaluation metric are in bold.}
\label{compare}
\end{table*}

\section{Analysis of the Language of Deceptive and Truthful Statements} \label{deceptivelanguage}

\subsection{Summary text statistics}

The summary statistics between truthful and deceptive statements using the \texttt{TEXTSTAT} python library are reported in Appendix~\ref{summary_stats}. We observe that the syllable \& lexicon count is significantly higher in deceptive statements (tested with an independent t-test).

\subsection{Vocabulary Uniqueness}

To examine the language of truthful and deceptive statements, we followed the steps by \citet{rissola2020beyond}, who used the Jaccard's index to measure the similarity between finite sample sets (here: truthful and deceptive statements). The Jaccard's index is calculated as follows: let $\mathcal{P}$ be the unique set of words obtained from truthful statements and $\mathcal{C}$ that from deceptive statements. The Jaccard's is calculated via the equation below:

\begin{equation}
    J(P,C) = |P \cap C|/|P \cup C|
\end{equation}

The Jaccard's index ranges from 0 to 1, where an index of 1 indicates that the sets completely intersect. As the value approaches 0, the sets have different elements. The sets $\mathcal{P}$ \& $\mathcal{C}$ intersect only to a low degree with Jaccard's index ranging from 0.33 to 0.36 (see Appendix~\ref{vocab_appendix} for full details), suggesting that people tend to use a different vocabulary for truthful and deceptive statements.

\begin{table}[hbt]
\scriptsize
\centering
\begin{tabular}{|c|c||c|c|}
\hline
\multicolumn{2}{|c||}{\textbf{Truthful Statements}}&\multicolumn{2}{c|}{\textbf{Deceptive Statements}}\\
\hline \hline
\textbf{LIWC} & \textbf{corr.} & \textbf{LIWC} & \textbf{corr.}\\ \hline
ingestion & 0.1695 & apostrophes & 0.1850 \\ \hline
\makecell[c]{biological \\processes} & 0.1397 & past focus & 0.1190 \\ \hline
\makecell[c]{analytical \\ thinking} & 0.1266 & reward & 0.1106 \\ \hline
numbers & 0.1228 & \makecell[c]{Word \\Count} & 0.0992 \\ \hline
leisure & 0.1209 & \makecell[c]{all \\ pronouns} & 0.0955\\ \hline
future focus & 0.1165 & \makecell[c]{personal \\pronoun} & 0.0915\\ \hline
\makecell[c]{dictionary \\ words} & 0.1106 & \makecell[c]{exclamation \\ marks} & 0.0906\\ \hline
article & 0.0930 & \makecell[c]{emotional \\ tone} & 0.0879\\ \hline
time & 0.0718 & \makecell[c]{present \\ focus} & 0.0789\\ \hline
affiliation & 0.0693 & \makecell[c]{positive \\ emotion} & 0.0785\\ \hline
home & 0.0665 & affect & 0.0749\\ \hline
- & - & \makecell[c]{all \\ punctuation} & 0.0724\\ \hline
- & - & adjective & 0.0662\\ \hline
- & - & \makecell[c]{commons \\ verbs} & 0.0657\\ \hline
- & - & \makecell[c]{3rd pers \\ singular} & 0.0627\\ \hline
\end{tabular}
\caption{Features associated with truthful and deceptive statements, sorted by point-biserial correlation. All correlations are significant at \textit{p} < 0.05 after Benjamini-Hochberg correction.}
\label{correlations}
\end{table}

\subsection{Feature Analysis - Correlation} \label{section_correlation}
To understand how individual features are related to each outcome class (truthful vs deceptive), we computed the point-biserial correlation of each LIWC\footnote{We provide examples of LIWC categories in Appendix~\ref{liwc_appendix}.} feature with the binary outcome class (truthful vs deceptive). Before calculating the correlation, we normalized the feature frequency for each document. The point-biserial correlation ranges from -1 to 1. Negative correlations were associated with truthful statements, whereas positive correlations were associated with deceptive statements. Thus, we considered the absolute values of the correlations. Table \ref{correlations} shows all significant correlations (\textit{p} < 0.05) after using the Benjamini-Hochberg alpha level correction procedure \citep{benjamini1995controlling}.

Table \ref{correlations} indicates  that the top-6 LIWC features which are  associated with truthful statements are ingestion, biological processes, analytical thinking, numbers, leisure, and future focus. This finding is illustrated in the following example (truthful text):  
\textit{'I will attend the party at 1pm at my friends Charlotte and David's home. There will be food available such as burgers and sausages and salads. I will sit in the shade if possible to avoid getting sunburnt'}

We further see that the majority of tokens correspond to the aforementioned LIWC features, i.e., ('will' --> 'future focus'), ('the' --> 'article'), ('food, burgers, sausages, salads' --> 'ingestion'), ('friends' --> 'affiliation').

The LIWC features correlated with deceptive statements are: apostrophes, past focus, reward, word count, all pronouns, personal pronoun, exclamation marks, emotional tone. The following example of a deceptive statement illustrates this:

\textit{'I recently got my motorcycle license, a few year after my friend who got his back when we were in uni. I haven't seen him since I graduated as he moved to Scotland and I work in London. We agreed to  go on a 2-day motorcycle holiday around the Brecon Becons as I have never seen them.'}

As one can easily observe, this deceptive statement includes verbs in past tense (LIWC category of past focus), apostrophes (haven't), all pronouns (my, his, we, I, we, them).

\subsection{Explainability-Error Analysis}

\begin{figure*}[t]
\centering
\includegraphics[width=1\textwidth]{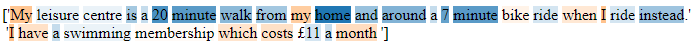}
\caption{Label: Truthful, Prediction: Truthful}
\label{lime14}
\end{figure*}

\begin{figure*}[t]
\centering
\includegraphics[width=1\textwidth]{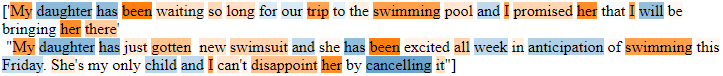}
\caption{Label: Deceptive, Prediction: Deceptive}
\label{lime71}
\end{figure*}

In this section we seek to further understand our best performing model from Section \ref{methods} (\textit{BERT+Co-Attention}). To do this, we employ LIME \citep{ribeiro2016should}, a model agnostic framework for interpretability. We experiment with 5000 samples and compare the findings of LIME with the results obtained in Section \ref{section_correlation}. The LIME results for (in)correct classifications of each class are illustrated in Figures \ref{lime14}-\ref{lime9}. Tokens in blue colour indicate those that were identified as truthful by LIME, whereas orange colour indicates tokens identified as deceptive. The examples can be summarised as follows:

\subsubsection{Correct classifications}

\begin{itemize}
    \item Truthful statements (Fig. \ref{lime14}): the model relied on personal pronouns as important for the deceptive class, while tokens corresponding to the LIWC categories leisure (leisure centre, walk, swimming), and home, were deemed indicative of the truthful class. Both results are in line with the findings presented in Section \ref{section_correlation}.
    \item Deceptive statements (Fig. \ref{lime71}): personal pronouns were correlated to the deceptive class, as was the LIWC category of past focus (promised, gotten), apostrophes (She's, can't), and affect (excited) were important for the deceptive class. However, tokens indicating future focus (will) seem important for the truthful class.
\end{itemize}

\subsubsection{Misclassifications}

\begin{figure*}[b]
\centering
\includegraphics[width=1\textwidth]{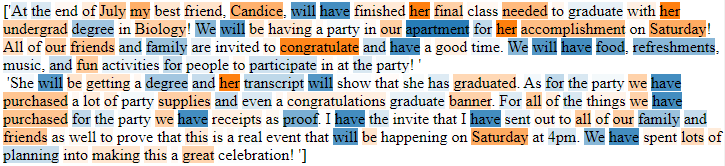}
\caption{Label: Truthful, Prediction: Deceptive}
\label{lime11}
\end{figure*}

\begin{figure*}[b]
\centering
\includegraphics[width=1\textwidth]{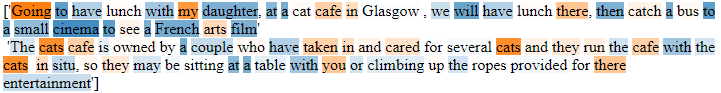}
\caption{Label: Deceptive, Prediction: Truthful}
\label{lime9}
\end{figure*}

\begin{itemize}
    \item Truthful statements (misclassified as deceptive) (Fig. \ref{lime11}): the model relies on tokens belonging to the LIWC categories of future focus (will), article (the), and ingestion (food), as being important for the truthful class. Similarly, it leans towards deceptive for the LIWC categories of personal pronouns, reward (congratulate, accomplishment), exclamation marks (!), past/present focus (graduated, purchased). However, the misclassification can be largely attributed to the majority of tokens belonging to the LIWC categories that are correlated with deceptive statements as described in Section \ref{section_correlation}.
    \item Deceptive statements (misclassified as truthful) (Fig. \ref{lime9}): similarly to the aforementioned example, this statement shows that the majority of tokens belong to the LIWC categories of future focus (will, may), article (a), leisure (cinema, entertainment) - all of which are correlated (for the whole corpus) with truthful statements. Those that correlated with the deceptive class (e.g., personal pronouns) are not frequent in this example, hence the risk for misclassification.
\end{itemize}

\section{Conclusions} \label{conclusion}

We face the challenge of differentiating truthful from deceptive texts in various aspects of our everyday life. In this paper, we have introduced several transformer-based network models and demonstrated that these models outperform other approaches reaching up to 70.61\% and 73.33\% in Accuracy and F1-score respectively. Furthermore, we have presented an in-depth analysis over the similarities and differences that can be observed between truthful and deceptive statements. We showed which LIWC categories correlate with truthful and deceptive texts. The findings revealed a mixed picture of similarities (e.g., number of sentences, etc.) and differences (e.g., different vocabulary, correlation with LIWC features). Finally, we have employed LIME to understand better how our best performing model classifies statements as truthful or deceptive. That analysis suggested that misclassifications are attributable to the presence of LIWC categories, which are significantly correlated with deceptive statements, in truthful statements and vice versa.

In the future, we aim to expand our approach to other datasets and domains and examine other transformer-based models (e.g., ALBERT, XLNet). A key challenge for applied verbal deception detection will lie in explainable methods and we aim to contribute to that task by further looking at statistical differences in language use as well as explainability techniques. We have shown how we can harness more advanced machine learning approaches without sacrificing explainability of these models.

\section*{Limitations}

\textbf{Hyperparameter Tuning:} In this work, we did not perform hyperparameter tuning. For instance, the value of \textit{k} in the co-attention mechanism, the number of units in the dense layers, the number of dense layers, the number of heads in the encoder part of the transformer, the learning rate have not been selected via the procedure of hyperparameter tuning. However, the procedure of fine-tuning leads to an increase in evaluation results. Therefore, the results of this study could be higher if the procedure of hyperparameter tuning had been adopted.

\noindent \textbf{Misclassifications:} Results showed that our model fails to correctly classify statements of the truthful class when they contain words significantly correlated with deceptive statements. Vice versa, some deceptive statements include words that belong to the LIWC categories presenting high correlation with truthful statements. Therefore, our model may need more contextualized information rather than relying on word frequencies alone to improve classification performance.

\noindent \textbf{Limitations of LIME:} We used LIME to explain the predictions of our best performing model. LIME constitutes an example of post-hoc approach, meaning that a second more explainable model must be trained. However, these learned surrogate models and the original models may rely on different mechanisms to make their predictions, leading to concerns about the fidelity of surrogate model-based approaches \citep{danilevsky-etal-2020-survey}. Therefore, future work could use interpretable (self-explaining) models to mitigate with the limitations of LIME.

\noindent \textbf{Training:} We trained our proposed approaches on two publicly available datasets for proving the robustness of our introduced approaches. This led to an increase of training time and use of GPU resources. In the future, we aim to adopt continual learning methods \citep{9349197} for saving time and adopting the same or even better performance on both datasets.

\section*{Ethical Considerations}

Our work brings with it a few ethical questions. First, if systems such as the current one (or related ones) were to be deployed for deception detection applications, we need to be aware of the considerable error rates that automated deception detection still involves. Of most concern here would be the false positives that incorrectly identify a statement (or person) as being deceptive (or a liar), which can create undesirable consequences (e.g., being marginalized). Second, especially in areas that touch law enforcement applications, we need to be cautious not to develop systems that are opaque in their working mechanisms, as they would violate the right to transparency. A prominent ethics framework (ALGOCARE) from the UK on data science applications in law enforcement \citep{doi:10.1080/13600834.2018.1458455} states clearly that a human must have the right to transparently trace how an automated decision was made. We hope that our work makes a step in that direction by looking beyond the classification performance of deep learning models.

\bibliography{anthology,custom}

\begin{thebibliography}{48}
\expandafter\ifx\csname natexlab\endcsname\relax\def\natexlab#1{#1}\fi

\bibitem[{Abadi et~al.(2016)Abadi, Agarwal, Barham, Brevdo, Chen, Citro,
  Corrado, Davis, Dean, Devin et~al.}]{abadi2016tensorflow}
Mart{\'\i}n Abadi, Ashish Agarwal, Paul Barham, Eugene Brevdo, Zhifeng Chen,
  Craig Citro, Greg~S Corrado, Andy Davis, Jeffrey Dean, Matthieu Devin, et~al.
  2016.
\newblock Tensorflow: Large-scale machine learning on heterogeneous distributed
  systems.
\newblock \emph{arXiv preprint arXiv:1603.04467}.

\bibitem[{Banerjee et~al.(2015)Banerjee, Chua, and
  Kim}]{10.1145/2701126.2701130}
Snehasish Banerjee, Alton Y.~K. Chua, and Jung-Jae Kim. 2015.
\newblock \href {https://doi.org/10.1145/2701126.2701130} {Using supervised
  learning to classify authentic and fake online reviews}.
\newblock In \emph{Proceedings of the 9th International Conference on
  Ubiquitous Information Management and Communication}, IMCOM '15, New York,
  NY, USA. Association for Computing Machinery.

\bibitem[{Benjamini and Hochberg(1995)}]{benjamini1995controlling}
Yoav Benjamini and Yosef Hochberg. 1995.
\newblock Controlling the false discovery rate: a practical and powerful
  approach to multiple testing.
\newblock \emph{Journal of the Royal statistical society: series B
  (Methodological)}, 57(1):289--300.

\bibitem[{Binder et~al.(2016)Binder, Bach, Montavon, M{\"u}ller, and
  Samek}]{binder2016layer}
Alexander Binder, Sebastian Bach, Gregoire Montavon, Klaus-Robert M{\"u}ller,
  and Wojciech Samek. 2016.
\newblock Layer-wise relevance propagation for deep neural network
  architectures.
\newblock In \emph{Information science and applications (ICISA) 2016}, pages
  913--922. Springer.

\bibitem[{Boyd et~al.(2022)Boyd, Ashokkumar, Seraj, and
  Pennebaker}]{boyd2022development}
Ryan~L Boyd, Ashwini Ashokkumar, Sarah Seraj, and James~W Pennebaker. 2022.
\newblock The development and psychometric properties of liwc-22.
\newblock \emph{Austin, TX: University of Texas at Austin}.

\bibitem[{Capuozzo et~al.(2020)Capuozzo, Lauriola, Strapparava, Aiolli, and
  Sartori}]{capuozzo-etal-2020-decop}
Pasquale Capuozzo, Ivano Lauriola, Carlo Strapparava, Fabio Aiolli, and
  Giuseppe Sartori. 2020.
\newblock \href {https://www.aclweb.org/anthology/2020.lrec-1.178} {{D}ec{O}p:
  A multilingual and multi-domain corpus for detecting deception in typed
  text}.
\newblock In \emph{Proceedings of the 12th Language Resources and Evaluation
  Conference}, pages 1423--1430, Marseille, France. European Language Resources
  Association.

\bibitem[{Chebbi and Jebara(2021)}]{chebbi2021deception}
Safa Chebbi and Sofia~Ben Jebara. 2021.
\newblock Deception detection using multimodal fusion approaches.
\newblock \emph{Multimedia Tools and Applications}, pages 1--30.

\bibitem[{Conroy et~al.(2015)Conroy, Rubin, and Chen}]{conroy2015automatic}
Nadia~K Conroy, Victoria~L Rubin, and Yimin Chen. 2015.
\newblock Automatic deception detection: Methods for finding fake news.
\newblock \emph{Proceedings of the Association for Information Science and
  Technology}, 52(1):1--4.

\bibitem[{Danilevsky et~al.(2020)Danilevsky, Qian, Aharonov, Katsis, Kawas, and
  Sen}]{danilevsky-etal-2020-survey}
Marina Danilevsky, Kun Qian, Ranit Aharonov, Yannis Katsis, Ban Kawas, and
  Prithviraj Sen. 2020.
\newblock \href {https://www.aclweb.org/anthology/2020.aacl-main.46} {A survey
  of the state of explainable {AI} for natural language processing}.
\newblock In \emph{Proceedings of the 1st Conference of the Asia-Pacific
  Chapter of the Association for Computational Linguistics and the 10th
  International Joint Conference on Natural Language Processing}, pages
  447--459, Suzhou, China. Association for Computational Linguistics.

\bibitem[{De~Lange et~al.(2022)De~Lange, Aljundi, Masana, Parisot, Jia,
  Leonardis, Slabaugh, and Tuytelaars}]{9349197}
Matthias De~Lange, Rahaf Aljundi, Marc Masana, Sarah Parisot, Xu~Jia, Ales
  Leonardis, Gregory Slabaugh, and Tinne Tuytelaars. 2022.
\newblock \href {https://doi.org/10.1109/TPAMI.2021.3057446} {A continual
  learning survey: Defying forgetting in classification tasks}.
\newblock \emph{IEEE Transactions on Pattern Analysis and Machine
  Intelligence}, 44(7):3366--3385.

\bibitem[{Devlin et~al.(2019)Devlin, Chang, Lee, and
  Toutanova}]{devlin2019bert}
Jacob Devlin, Ming-Wei Chang, Kenton Lee, and Kristina Toutanova. 2019.
\newblock \href {http://arxiv.org/abs/1810.04805} {Bert: Pre-training of deep
  bidirectional transformers for language understanding}.

\bibitem[{Fitzpatrick et~al.(2015)Fitzpatrick, Bachenko, and
  Fornaciari}]{fitzpatrick2015automatic}
Eileen Fitzpatrick, Joan Bachenko, and Tommaso Fornaciari. 2015.
\newblock Automatic detection of verbal deception.
\newblock \emph{Synthesis Lectures on Human Language Technologies},
  8(3):1--119.

\bibitem[{Fontanarava et~al.(2017)Fontanarava, Pasi, and Viviani}]{8259828}
Julien Fontanarava, Gabriella Pasi, and Marco Viviani. 2017.
\newblock \href {https://doi.org/10.1109/DSAA.2017.51} {Feature analysis for
  fake review detection through supervised classification}.
\newblock In \emph{2017 IEEE International Conference on Data Science and
  Advanced Analytics (DSAA)}, pages 658--666.

\bibitem[{Forman et~al.(2003)}]{forman2003extensive}
George Forman et~al. 2003.
\newblock An extensive empirical study of feature selection metrics for text
  classification.
\newblock \emph{J. Mach. Learn. Res.}, 3(Mar):1289--1305.

\bibitem[{Fornaciari et~al.(2021)Fornaciari, Bianchi, Poesio, and
  Hovy}]{fornaciari-etal-2021-bertective}
Tommaso Fornaciari, Federico Bianchi, Massimo Poesio, and Dirk Hovy. 2021.
\newblock \href {https://www.aclweb.org/anthology/2021.eacl-main.232}
  {{BERT}ective: Language models and contextual information for deception
  detection}.
\newblock In \emph{Proceedings of the 16th Conference of the European Chapter
  of the Association for Computational Linguistics: Main Volume}, pages
  2699--2708, Online. Association for Computational Linguistics.

\bibitem[{Fornaciari and Poesio(2012)}]{fornaciari-poesio-2012-decour}
Tommaso Fornaciari and Massimo Poesio. 2012.
\newblock \href
  {http://www.lrec-conf.org/proceedings/lrec2012/pdf/377_Paper.pdf}
  {{D}e{C}our: a corpus of {DE}ceptive statements in {I}talian {COUR}ts}.
\newblock In \emph{Proceedings of the Eighth International Conference on
  Language Resources and Evaluation ({LREC}'12)}, pages 1585--1590, Istanbul,
  Turkey. European Language Resources Association (ELRA).

\bibitem[{Fornaciari and Poesio(2013)}]{fornaciari2013automatic}
Tommaso Fornaciari and Massimo Poesio. 2013.
\newblock Automatic deception detection in italian court cases.
\newblock \emph{Artificial intelligence and law}, 21(3):303--340.

\bibitem[{Girgis et~al.(2018)Girgis, Amer, and Gadallah}]{girgis2018deep}
Sherry Girgis, Eslam Amer, and Mahmoud Gadallah. 2018.
\newblock Deep learning algorithms for detecting fake news in online text.
\newblock In \emph{2018 13th International Conference on Computer Engineering
  and Systems (ICCES)}, pages 93--97. IEEE.

\bibitem[{Hajek et~al.(2020)Hajek, Barushka, and Munk}]{hajek2020fake}
Petr Hajek, Aliaksandr Barushka, and Michal Munk. 2020.
\newblock Fake consumer review detection using deep neural networks integrating
  word embeddings and emotion mining.
\newblock \emph{Neural Computing and Applications}, 32(23):17259--17274.

\bibitem[{{Jin} et~al.(2019){Jin}, {Wei}, {Du}, {Xue}, and
  {Ren}}]{2019arXiv191106194J}
Xisen {Jin}, Zhongyu {Wei}, Junyi {Du}, Xiangyang {Xue}, and Xiang {Ren}. 2019.
\newblock \href {http://arxiv.org/abs/1911.06194} {{Towards Hierarchical
  Importance Attribution: Explaining Compositional Semantics for Neural
  Sequence Models}}.
\newblock \emph{arXiv e-prints}, page arXiv:1911.06194.

\bibitem[{Kao et~al.(2020)Kao, Chen, Tzeng, Chen, Shmueli, and
  Ku}]{kao2020detecting}
Yi-Ying Kao, Po-Han Chen, Chun-Chiao Tzeng, Zi-Yuan Chen, Boaz Shmueli, and
  Lun-Wei Ku. 2020.
\newblock Detecting deceptive language in crime interrogation.
\newblock In \emph{International Conference on Human-Computer Interaction},
  pages 80--90. Springer.

\bibitem[{Kennedy et~al.(2019)Kennedy, Walsh, Sloka, McCarren, and
  Foster}]{kennedy-etal-2019-fact}
Stefan Kennedy, Niall Walsh, Kirils Sloka, Andrew McCarren, and Jennifer
  Foster. 2019.
\newblock \href {https://doi.org/10.18653/v1/P19-2048} {Fact or factitious?
  contextualized opinion spam detection}.
\newblock In \emph{Proceedings of the 57th Annual Meeting of the Association
  for Computational Linguistics: Student Research Workshop}, pages 344--350,
  Florence, Italy. Association for Computational Linguistics.

\bibitem[{Khan et~al.(2021)Khan, Crockett, O'Shea, Hussain, and
  Khan}]{KHAN2021114341}
Wasiq Khan, Keeley Crockett, James O'Shea, Abir Hussain, and Bilal~M. Khan.
  2021.
\newblock \href {https://doi.org/https://doi.org/10.1016/j.eswa.2020.114341}
  {Deception in the eyes of deceiver: A computer vision and machine learning
  based automated deception detection}.
\newblock \emph{Expert Systems with Applications}, 169:114341.

\bibitem[{Kleinberg et~al.(2018)Kleinberg, Mozes, Arntz, and
  Verschuere}]{kleinberg2018using}
Bennett Kleinberg, Maximilian Mozes, Arnoud Arntz, and Bruno Verschuere. 2018.
\newblock Using named entities for computer-automated verbal deception
  detection.
\newblock \emph{Journal of forensic sciences}, 63(3):714--723.

\bibitem[{Kleinberg et~al.(2017)Kleinberg, Nahari, Arntz, and
  Verschuere}]{kleinberg2017investigation}
Bennett Kleinberg, Galit Nahari, Arnoud Arntz, and Bruno Verschuere. 2017.
\newblock An investigation on the detectability of deceptive intent about
  flying through verbal deception detection.
\newblock \emph{Collabra: Psychology}, 3(1).

\bibitem[{Kleinberg and Verschuere(2021)}]{KLEINBERG2021103250}
Bennett Kleinberg and Bruno Verschuere. 2021.
\newblock \href {https://doi.org/https://doi.org/10.1016/j.actpsy.2020.103250}
  {How humans impair automated deception detection performance}.
\newblock \emph{Acta Psychologica}, 213:103250.

\bibitem[{Li et~al.(2015)Li, Chen, Hovy, and Jurafsky}]{li2015visualizing}
Jiwei Li, Xinlei Chen, Eduard Hovy, and Dan Jurafsky. 2015.
\newblock Visualizing and understanding neural models in nlp.
\newblock \emph{arXiv preprint arXiv:1506.01066}.

\bibitem[{Liu et~al.(2019)Liu, Ott, Goyal, Du, Joshi, Chen, Levy, Lewis,
  Zettlemoyer, and Stoyanov}]{https://doi.org/10.48550/arxiv.1907.11692}
Yinhan Liu, Myle Ott, Naman Goyal, Jingfei Du, Mandar Joshi, Danqi Chen, Omer
  Levy, Mike Lewis, Luke Zettlemoyer, and Veselin Stoyanov. 2019.
\newblock \href {https://doi.org/10.48550/ARXIV.1907.11692} {Roberta: A
  robustly optimized bert pretraining approach}.

\bibitem[{Lu et~al.(2016)Lu, Yang, Batra, and Parikh}]{NIPS2016_9dcb88e0}
Jiasen Lu, Jianwei Yang, Dhruv Batra, and Devi Parikh. 2016.
\newblock \href
  {https://proceedings.neurips.cc/paper/2016/file/9dcb88e0137649590b755372b040afad-Paper.pdf}
  {Hierarchical question-image co-attention for visual question answering}.
\newblock In \emph{Advances in Neural Information Processing Systems},
  volume~29. Curran Associates, Inc.

\bibitem[{Nahari et~al.(2019)Nahari, Ashkenazi, Fisher, Granhag, Hershkowitz,
  Masip, Meijer, Nisin, Sarid, Taylor et~al.}]{nahari2019language}
Galit Nahari, Tzachi Ashkenazi, Ronald~P Fisher, P{\"a}r-Anders Granhag, Irit
  Hershkowitz, Jaume Masip, Ewout~H Meijer, Zvi Nisin, Nadav Sarid, Paul~J
  Taylor, et~al. 2019.
\newblock ‘language of lies’: Urgent issues and prospects in verbal lie
  detection research.
\newblock \emph{Legal and Criminological Psychology}, 24(1):1--23.

\bibitem[{Nguyen et~al.(2018)Nguyen, Kharosekar, Lease, and
  Wallace}]{Nguyen_Kharosekar_Lease_Wallace_2018}
An~Nguyen, Aditya Kharosekar, Matthew Lease, and Byron Wallace. 2018.
\newblock \href {https://doi.org/10.1609/aaai.v32i1.11487} {An interpretable
  joint graphical model for fact-checking from crowds}.
\newblock \emph{Proceedings of the AAAI Conference on Artificial Intelligence},
  32(1).

\bibitem[{Oswald et~al.(2018)Oswald, Grace, Urwin, and
  Barnes}]{doi:10.1080/13600834.2018.1458455}
Marion Oswald, Jamie Grace, Sheena Urwin, and Geoffrey~C. Barnes. 2018.
\newblock \href {https://doi.org/10.1080/13600834.2018.1458455} {Algorithmic
  risk assessment policing models: lessons from the durham hart model and
  ‘experimental’ proportionality}.
\newblock \emph{Information \& Communications Technology Law}, 27(2):223--250.

\bibitem[{Ott et~al.(2011)Ott, Choi, Cardie, and
  Hancock}]{ott-etal-2011-finding}
Myle Ott, Yejin Choi, Claire Cardie, and Jeffrey~T. Hancock. 2011.
\newblock \href {https://www.aclweb.org/anthology/P11-1032} {Finding deceptive
  opinion spam by any stretch of the imagination}.
\newblock In \emph{Proceedings of the 49th Annual Meeting of the Association
  for Computational Linguistics: Human Language Technologies}, pages 309--319,
  Portland, Oregon, USA. Association for Computational Linguistics.

\bibitem[{Pennebaker et~al.(2015)Pennebaker, Boyd, Jordan, and
  Blackburn}]{pennebaker2015development}
James~W Pennebaker, Ryan~L Boyd, Kayla Jordan, and Kate Blackburn. 2015.
\newblock The development and psychometric properties of liwc2015.
\newblock Technical report.

\bibitem[{P{\'e}rez-Rosas et~al.(2015)P{\'e}rez-Rosas, Abouelenien, Mihalcea,
  Xiao, Linton, and Burzo}]{perez-rosas-etal-2015-verbal}
Ver{\'o}nica P{\'e}rez-Rosas, Mohamed Abouelenien, Rada Mihalcea, Yao Xiao,
  CJ~Linton, and Mihai Burzo. 2015.
\newblock \href {https://doi.org/10.18653/v1/D15-1281} {Verbal and nonverbal
  clues for real-life deception detection}.
\newblock In \emph{Proceedings of the 2015 Conference on Empirical Methods in
  Natural Language Processing}, pages 2336--2346, Lisbon, Portugal. Association
  for Computational Linguistics.

\bibitem[{P{\'e}rez-Rosas et~al.(2018)P{\'e}rez-Rosas, Kleinberg, Lefevre, and
  Mihalcea}]{perez-rosas-etal-2018-automatic}
Ver{\'o}nica P{\'e}rez-Rosas, Bennett Kleinberg, Alexandra Lefevre, and Rada
  Mihalcea. 2018.
\newblock \href {https://aclanthology.org/C18-1287} {Automatic detection of
  fake news}.
\newblock In \emph{Proceedings of the 27th International Conference on
  Computational Linguistics}, pages 3391--3401, Santa Fe, New Mexico, USA.
  Association for Computational Linguistics.

\bibitem[{P{\'e}rez-Rosas and
  Mihalcea(2015)}]{perez-rosas-mihalcea-2015-experiments}
Ver{\'o}nica P{\'e}rez-Rosas and Rada Mihalcea. 2015.
\newblock \href {https://doi.org/10.18653/v1/D15-1133} {Experiments in open
  domain deception detection}.
\newblock In \emph{Proceedings of the 2015 Conference on Empirical Methods in
  Natural Language Processing}, pages 1120--1125, Lisbon, Portugal. Association
  for Computational Linguistics.

\bibitem[{Ribeiro et~al.(2016)Ribeiro, Singh, and Guestrin}]{ribeiro2016should}
Marco~Tulio Ribeiro, Sameer Singh, and Carlos Guestrin. 2016.
\newblock " why should i trust you?" explaining the predictions of any
  classifier.
\newblock In \emph{Proceedings of the 22nd ACM SIGKDD international conference
  on knowledge discovery and data mining}, pages 1135--1144.

\bibitem[{R{\'\i}ssola et~al.(2020)R{\'\i}ssola, Aliannejadi, and
  Crestani}]{rissola2020beyond}
Esteban~Andr{\'e}s R{\'\i}ssola, Mohammad Aliannejadi, and Fabio Crestani.
  2020.
\newblock Beyond modelling: understanding mental disorders in online social
  media.
\newblock In \emph{European Conference on Information Retrieval}, pages
  296--310. Springer.

\bibitem[{Rubin et~al.(2015)Rubin, Chen, and Conroy}]{rubin2015deception}
Victoria~L Rubin, Yimin Chen, and Nadia~K Conroy. 2015.
\newblock Deception detection for news: three types of fakes.
\newblock \emph{Proceedings of the Association for Information Science and
  Technology}, 52(1):1--4.

\bibitem[{Simonyan et~al.(2014)Simonyan, Vedaldi, and
  Zisserman}]{simonyan2014deep}
Karen Simonyan, Andrea Vedaldi, and Andrew Zisserman. 2014.
\newblock \href {http://arxiv.org/abs/1312.6034} {Deep inside convolutional
  networks: Visualising image classification models and saliency maps}.

\bibitem[{Soldner et~al.(2019)Soldner, P{\'e}rez-Rosas, and
  Mihalcea}]{soldner-etal-2019-box}
Felix Soldner, Ver{\'o}nica P{\'e}rez-Rosas, and Rada Mihalcea. 2019.
\newblock \href {https://doi.org/10.18653/v1/N19-1175} {Box of lies: Multimodal
  deception detection in dialogues}.
\newblock In \emph{Proceedings of the 2019 Conference of the North {A}merican
  Chapter of the Association for Computational Linguistics: Human Language
  Technologies, Volume 1 (Long and Short Papers)}, pages 1768--1777,
  Minneapolis, Minnesota. Association for Computational Linguistics.

\bibitem[{Tausczik and Pennebaker(2010)}]{Tausczik2010ThePM}
Yla~R. Tausczik and James~W. Pennebaker. 2010.
\newblock The psychological meaning of words: Liwc and computerized text
  analysis methods.
\newblock \emph{Journal of Language and Social Psychology}, 29:24 -- 54.

\bibitem[{Tomas et~al.(2022)Tomas, Dodier, and
  Demarchi}]{tomas2022computational}
Fr{\'e}d{\'e}ric Tomas, Olivier Dodier, and Samuel Demarchi. 2022.
\newblock Computational measures of deceptive language: prospects and issues.
\newblock \emph{Frontiers in Communication}, 7:792378.

\bibitem[{Vaswani et~al.(2017)Vaswani, Shazeer, Parmar, Uszkoreit, Jones,
  Gomez, Kaiser, and Polosukhin}]{NIPS2017_3f5ee243}
Ashish Vaswani, Noam Shazeer, Niki Parmar, Jakob Uszkoreit, Llion Jones,
  Aidan~N Gomez, \L~ukasz Kaiser, and Illia Polosukhin. 2017.
\newblock \href
  {https://proceedings.neurips.cc/paper/2017/file/3f5ee243547dee91fbd053c1c4a845aa-Paper.pdf}
  {Attention is all you need}.
\newblock In \emph{Advances in Neural Information Processing Systems},
  volume~30. Curran Associates, Inc.

\bibitem[{Vrij et~al.(2010)Vrij, Granhag, and Porter}]{vrij2010pitfalls}
Aldert Vrij, P{\"a}r~Anders Granhag, and Stephen Porter. 2010.
\newblock Pitfalls and opportunities in nonverbal and verbal lie detection.
\newblock \emph{Psychological science in the public interest}, 11(3):89--121.

\bibitem[{Wolf et~al.(2020)Wolf, Debut, Sanh, Chaumond, Delangue, Moi, Cistac,
  Rault, Louf, Funtowicz, Davison, Shleifer, von Platen, Ma, Jernite, Plu, Xu,
  Scao, Gugger, Drame, Lhoest, and Rush}]{wolf-etal-2020-transformers}
Thomas Wolf, Lysandre Debut, Victor Sanh, Julien Chaumond, Clement Delangue,
  Anthony Moi, Pierric Cistac, Tim Rault, Rémi Louf, Morgan Funtowicz, Joe
  Davison, Sam Shleifer, Patrick von Platen, Clara Ma, Yacine Jernite, Julien
  Plu, Canwen Xu, Teven~Le Scao, Sylvain Gugger, Mariama Drame, Quentin Lhoest,
  and Alexander~M. Rush. 2020.
\newblock \href {https://www.aclweb.org/anthology/2020.emnlp-demos.6}
  {Transformers: State-of-the-art natural language processing}.
\newblock In \emph{Proceedings of the 2020 Conference on Empirical Methods in
  Natural Language Processing: System Demonstrations}, pages 38--45, Online.
  Association for Computational Linguistics.

\bibitem[{Wu et~al.(2018)Wu, Singh, Davis, and
  Subrahmanian}]{Wu_Singh_Davis_Subrahmanian_2018}
Zhe Wu, Bharat Singh, Larry Davis, and V.~Subrahmanian. 2018.
\newblock \href {https://ojs.aaai.org/index.php/AAAI/article/view/11502}
  {Deception detection in videos}.
\newblock \emph{Proceedings of the AAAI Conference on Artificial Intelligence},
  32(1).

\end{thebibliography}
\bibliographystyle{acl_natbib}

\appendix

\section{Results of RoBERTa} \label{results_roberta}

In Table \ref{compare_roberta}, the results of our proposed approaches (by employing RoBERTa) are reported.

\begin{table*}[hbt]
\scriptsize
\centering
\begin{tabular}{lcccccc}
\toprule
\multicolumn{1}{l}{}&\multicolumn{6}{c}{\textbf{Evaluation metrics}}\\
\cline{2-7} 
\multicolumn{1}{l}{\textbf{Architecture}}&\textbf{Precision}&\textbf{Recall}&\textbf{F1-score}&\textbf{Accuracy}&\textbf{AUROC}&\textbf{Specificity}\\
\midrule
\textit{\small{RoBERTa + Dense Layers}} & 63.56 & 41.99 & 47.64 & 56.15 & 61.24 & 71.74 \\
& $\pm$6.07 & $\pm$17.93 & $\pm$13.85 & $\pm$3.99 & $\pm$3.94 & $\pm$16.20\\ \hline
\textit{\small{RoBERTa + MultiHead Attention}} & 66.75 & 47.84 & 54.99 & 60.01 & 66.35 & \textbf{73.42} \\
& $\pm$4.75 & $\pm$10.64 & $\pm$7.29 & $\pm$3.60 & $\pm$4.34 & $\pm$7.98\\ \hline
\textit{\small{RoBERTa + Transformers}} & 67.58 & 52.26 & 57.53 & 61.90 & 68.21 & 72.40\\
& $\pm$3.33 & $\pm$15.33 & $\pm$10.79 & $\pm$4.52 & $\pm$4.92 & $\pm$9.26 \\ \hline
\textit{\small{RoBERTa + Co-Attention}}& 69.31 & 76.89 & \textbf{72.44} & 69.70 & 76.68 & 61.84  \\
& $\pm$3.94 & $\pm$8.60 & $\pm$2.81 & $\pm$2.25 & $\pm$2.12 & $\pm$10.57 \\ \hline
\textit{\small{RoBERTa + Co-Attention + LIWC}} & \textbf{70.47} & 73.40 & 71.22 & 69.31 & \textbf{76.91} & 64.79 \\
& $\pm$5.39 & $\pm$9.47 & $\pm$3.18 & $\pm$2.83 & $\pm$2.42 & $\pm$12.94\\ \hline
\textit{\small{RoBERTa + Transformers + Co-Attention}} & 69.55 & \textbf{75.70} & 72.38 & \textbf{69.86} & 76.67 & 63.50 \\
& $\pm$3.17 & $\pm$4.14 & $\pm$2.11 & $\pm$2.34 & $\pm$2.52 & $\pm$5.96\\ 
\bottomrule
\end{tabular}
\caption{Performance of proposed deception detection models (RoBERTa) using the dataset \citep{KLEINBERG2021103250}. Reported values are means $\pm$ standard deviation. Best results per evaluation metric are in bold.}
\label{compare_roberta}
\end{table*}

\section{Results of the introduced approaches on the Open Domain Deception Dataset} \label{open_domain_results}

Table \ref{comparee} shows that BERT+Co-Attention constitutes our best performing model, since it outperforms the other proposed models in terms of Accuracy, Recall, F1-score, and AUROC. Although it performs worse in Precision, it surpasses all models in the F1-score, which constitutes the weighted average of Precision and Recall. Moreover, it surpasses all other models models in Accuracy, in F1-score, in AUROC, and in Recall by 1.01-12.45, 2.23-20.67, 1.59-16.63, and 5.38-27.12 points respectively. However, it performs worse than \textit{Feature Set: POS} by 1.09 points. 

\begin{table*}[hbt]
\scriptsize
\centering
\begin{tabular}{lcccccc}
\toprule
\multicolumn{1}{l}{}&\multicolumn{6}{c}{\textbf{Evaluation metrics}}\\
\cline{2-7} 
\multicolumn{1}{l}{\textbf{Architecture}}&\textbf{Precision}&\textbf{Recall}&\textbf{F1-score}&\textbf{Accuracy}&\textbf{AUROC}&\textbf{Specificity}\\
\hline
\textit{\small{Feature Set: POS}} & - & - & - & \textbf{69.50} & - & -\\
\hline \hline
\textit{\small{BERT + Dense Layers}} & 60.42 & 57.10 & 57.41 & 59.27 & 62.99 & 61.53 \\
& $\pm$5.37 & $\pm$13.96 & $\pm$7.11 & $\pm$3.53 & $\pm$3.45 & $\pm$15.05\\ \hline
\textit{\small{BERT + MultiHead Attention}} & 58.83 & 43.59 & 48.10  & 55.96 & 59.61 & 68.30 \\
& $\pm$6.80 & $\pm$16.60 & $\pm$10.95 & $\pm$3.80 & $\pm$4.88 & $\pm$17.07\\ \hline
\textit{\small{BERT + Transformers}} & \textbf{68.53} & 65.02 & 66.54 & 67.40 & 74.65 & \textbf{69.75} \\
& $\pm$3.71 & $\pm$6.26 & $\pm$4.02 & $\pm$2.99 & $\pm$2.93 & $\pm$5.79\\ \hline
\textit{\small{BERT + Co-Attention}}& 68.52 & \textbf{70.71} & \textbf{68.77} & 68.41 & \textbf{76.24} & 66.12  \\
& $\pm$5.19 & $\pm$10.77 & $\pm$4.64 & $\pm$3.03 & $\pm$2.74 & $\pm$11.67 \\ \hline
\textit{\small{BERT + Transformers + Co-Attention}} & 67.16 & 65.33 & 66.14 & 66.71 & 73.18 & 68.07 \\
& $\pm$2.06 & $\pm$5.37 & $\pm$3.44 & $\pm$2.44 & $\pm$3.03 & $\pm$3.02\\ 
\bottomrule
\end{tabular}
\caption{Performance comparison between previous \citep{perez-rosas-mihalcea-2015-experiments} and proposed deception detection models. Reported values are means $\pm$ standard deviation. Best results per evaluation metric are in bold.}
\label{comparee}
\end{table*}

\section{Summary Text Statistics} \label{summary_stats}

Table \ref{textstat} reports the results of the \texttt{TEXTSTAT} library.

\begin{table*}[hbt]
\scriptsize
\centering
\begin{tabular}{ccccccc}
\toprule
\multicolumn{1}{c}{} & \multicolumn{2}{c}{Q1} & \multicolumn{2}{c}{Q2} & \multicolumn{2}{c}{Q1 \& Q2} \\ \cline{2-7}
\textbf{Metric} & T & D & T & D & T & D \\ \midrule
\small Syllable Count & 59.9 $\pm$ 36.0$\dagger$ & 67.5 $\pm$ 39.5 & 45.3 $\pm$ 30.7$\dagger$ & 50.7 $\pm$ 32.2 & 104.6 $\pm$ 57.1$\dagger$ & 117.6 $\pm$ 59.9\\
\small Lexicon Count & 46.2 $\pm$ 27.7$\dagger$ & 52 $\pm$ 30.4 & 34.5 $\pm$ 23.4$\dagger$ & 38.9 $\pm$ 24.4 & 79.9 $\pm$ 43.3$\dagger$ & 90.1 $\pm$ 45.7 \\
\small Sentence Count & 2.3 $\pm$ 1.6 & 2.4 $\pm$ 1.5 & 1.9 $\pm$ 1.3 & 2.0 $\pm$ 1.2 & 3.4 $\pm$ 2.5$\dagger$ & 3.6 $\pm$ 2.4 \\
\small Difficult Words & 4.9 $\pm$ 3.8$\dagger$ & 5.4 $\pm$ 4.0 & 4.1 $\pm$ 3.4$\dagger$ & 4.6 $\pm$ 3.7 & 8.5 $\pm$ 5.8$\dagger$ & 9.2 $\pm$ 6.0\\ \bottomrule
\end{tabular}
\caption{Average + standard deviation metrics per statement(s). $\dagger$ indicates statistical significance between truthful and deceptive statements. All differences are significant at $p<0.05$ after Benjamini-Hochberg correction.}
\label{textstat}
\end{table*}

\section{Vocabulary Uniqueness} \label{vocab_appendix}

Table \ref{jaccard_index} reports the results of Jaccard Index between truthful and deceptive statements.

\begin{table*}[hbt]
\centering
\begin{tabular}{cc}
\toprule
\textbf{Jaccard Index between statements} & \textbf{Result} \\ \midrule
 \makecell{\textit{J}($\mathcal{P}$= truthful $Q_1$ \& $Q_2$, \\$\mathcal{C}$=deceptive $Q_1$ \& $Q_2$)} & 0.3618 \\ \hline
 \textit{J}($\mathcal{P}$= truthful $Q_1$, $\mathcal{C}$=deceptive $Q_1$) & 0.3548 \\ \hline
 \textit{J}($\mathcal{P}$= truthful $Q_2$, $\mathcal{C}$=deceptive $Q_2$) & 0.3341 \\ \bottomrule
\end{tabular}
\caption{Jaccard Index between truthful and deceptive statements}
\label{jaccard_index}
\end{table*}

\section{LIWC Categories} \label{liwc_appendix}

Table \ref{liwc_categories} provides an overview of the LIWC features along with some specific examples.

\begin{table*}[hbt]
\centering
\begin{tabular}{ll}
\toprule
\textbf{LIWC categories} & \textbf{Examples} \\ \midrule
ingestion & dish, eat, pizza \\ \hline
biological processes  & eat, blood, pain \\ \hline
analytical thinking  &  \\ \hline
numbers  &  second, thousand\\ \hline
leisure  &  cook, chat, movie\\ \hline
future focus  & may, will, soon \\ \hline
dictionary words  &  \\ \hline
article  & a, an, the \\ \hline
time  &  end, until, season\\ \hline
affiliation  & ally, friend, social \\ \hline
home  &  kitchen, landlord \\ \hline
apostrophes  & ' \\ \hline
past focus  & ago, did, talked  \\ \hline
reward  & take, prize, benefit  \\ \hline
word count  &  \\ \hline
all pronouns  & I, them, itself \\ \hline
personal pronoun & I, them, her \\ \hline
exclamation marks   &  !\\ \hline
emotional tone  &  \\ \hline
present focus   & today, is, now\\ \hline
positive emotion  & love, nice, sweet \\ \hline
affect  & happy, cried  \\ \hline
all punctuation  & periods, colons, etc. \\ \hline
adjective  & ree, happy, long  \\ \hline
common verbs  & eat, come, carry \\ \hline
3rd pers singular & she, he \\ \bottomrule
\end{tabular}
\caption{LIWC categories}
\label{liwc_categories}
\end{table*}

\end{document}